\documentclass[10pt,twocolumn,letterpaper]{article}

\usepackage{cvpr}              

\usepackage{graphicx}
\usepackage{amsmath}
\usepackage{amssymb}
\usepackage{booktabs}

\usepackage{multirow}

\usepackage[pagebackref,breaklinks,colorlinks]{hyperref}

\usepackage[capitalize]{cleveref}
\crefname{section}{Sec.}{Secs.}
\Crefname{section}{Section}{Sections}
\Crefname{table}{Table}{Tables}
\crefname{table}{Tab.}{Tabs.}

\def\cvprPaperID{4082} 
\def\confName{CVPR}
\def\confYear{2022}

\begin{document}

\title{Feature-Filter: Detecting Adversarial Examples through Filtering out Recessive Features}

\author{Hui Liu \; Bo Zhao \; Minzhi Ji \; Yuefeng Peng \; Jiabao Guo \\
School of Cyber Science and Engineering, Wuhan University\\
Wuhan, 430072 China\\
{\tt\small \{liuh824, zhaobo, jiminzhi, yuefengpeng, garbo\_guo\}@whu.edu.cn }
\and
Peng Liu\\
College of Information Sciences and Technology Pennsylvania State University\\
PA, 16801 US\\
{\tt\small pliu@ist.psu.edu}
}
\maketitle


\begin{abstract}

Deep neural networks (DNNs) are under threat from adversarial example attacks. The adversary can easily change the outputs of DNNs by adding small well-designed perturbations to inputs. Adversarial example detection is a fundamental work for robust DNNs-based service. Adversarial examples show the difference between humans and DNNs in image recognition. From a human-centric perspective, image features could be divided into dominant features comprehensible to humans, and recessive features incomprehensible to humans, yet exploited by DNNs. In this paper, we reveal that imperceptible adversarial examples are the product of recessive features misleading neural networks, and the adversarial attack enriches these recessive features. The imperceptibility of the adversarial examples indicates that the perturbations enrich recessive features, yet hardly affect dominant features. Therefore, adversarial examples are sensitive to filtering out recessive features, while benign examples are immune to such operation. Inspired by this idea, we propose a label-only adversarial detection approach that is referred to as feature-filter. Feature-filter utilizes discrete cosine transform (DCT) to approximately separate recessive features from dominant features and gets a filtered image. A comprehensive user study demonstrates that the DCT-based filter can reliably filter out recessive features from the test image. By only comparing DNN's prediction labels on the input and its filtered version, feature-filter can real-time detect imperceptible adversarial examples at high accuracy and few false positives.

\end{abstract}

\section{Introduction}
\label{sec:intro}

Deep neural networks (DNNs) \cite{HeK, HuangG, Karen, HuJ} achieve state-of-the-art performance on many artificial intelligence tasks, including safety-critical systems like facial biometric systems \cite{BlancoG}, intrusion detection system \cite{MirskyY}, etc. However, a large number of studies have shown that attackers can easily change the outputs of DNNs by the inputs with well-designed perturbations (adversarial examples) \cite{Szegedy}. The malicious action is called ``adversarial attack"  \cite{Goodfellow, YuanX, Dezfooli, Universal, DongJ, Bhattad, SuJ, Laidlaw}.

Adversaries could modify any pixel value in the image to generate adversarial examples. In the physical world, however, they don't want modifications that can easily irritate the human eye. For example, autonomous vehicles deploy extensive neural networks to construct their perceptual systems \cite{Chiaroni, XuX}. Adversaries could try to affect the decision-making of perceptual systems by using adversarial traffic signs \cite{Lovisotto}. If these modifications in traffic signs can be easily detected by the human eye, the traffic police would replace them in time and the adversarial attack would fail before it had even begun. Therefore, in order to achieve this goal, adversaries must ensure that these modifications are imperceptible to the human eye. This constraint of imperceptibility is widespread in real-world scenarios.

The phenomenon of adversarial examples has attracted a great number of interests in the research community in recent years. To hardening DNN systems, A wide range of proactive defense approaches \cite{PeiK, Shaham, Kurakin, Tram, Papernot} have been proposed in image classification, but those defenses can be evaded by emerging attack approaches \cite{Carlini}. If there are no known intrinsic properties that distinguish adversarial examples from benign examples, these proactive adversarial defenses are extremely challenging \cite{CohenG}.

As a fundamental work for robust DNNs, detecting adversarial examples may be as important as adversarial defenses. Adversarial detection aims to distinguish adversarial examples from benign ones by intrinsic properties. To discover these intrinsic properties, let us analyze the reasons for the existence of imperceptible adversarial examples in the image recognition.

There has been a debate on why the adversarial examples exist. Previous work has proposed several explanations for the phenomenon of adversarial examples, ranging from finite-sample overfitting to high-dimensional statistical properties \cite{Bubeck, Shafahi, Mahloujifar}. However, Andrew et al. \cite{Ilyas} argued these theories were often unable to fully capture behaviors, and first categorized image features from a human-centric perspective. Inspired by this study, we observe that adversarial examples show the difference between humans and DNNs in the utilization of input features. For accurate classification, DNNs make full use of the input features, of which spaces are often unnecessarily large for humans. The human eyes recognize images only according to those features that are comprehensible to humans. As a human-centric phenomenon, the input features could be divided into two categories: dominant features comprehensible to humans, and recessive features incomprehensible to humans. When adversarial perturbations enough enrich recessive features and hardly lose dominant features, they would mislead the output of DNNs and not result in noticeable artifacts to the human eyes. Thus, we claim that the existence of imperceptible adversarial examples is due to the neural network's use of these recessive features.

\begin{figure}[htb]
\centering
\includegraphics[width=0.65\columnwidth]{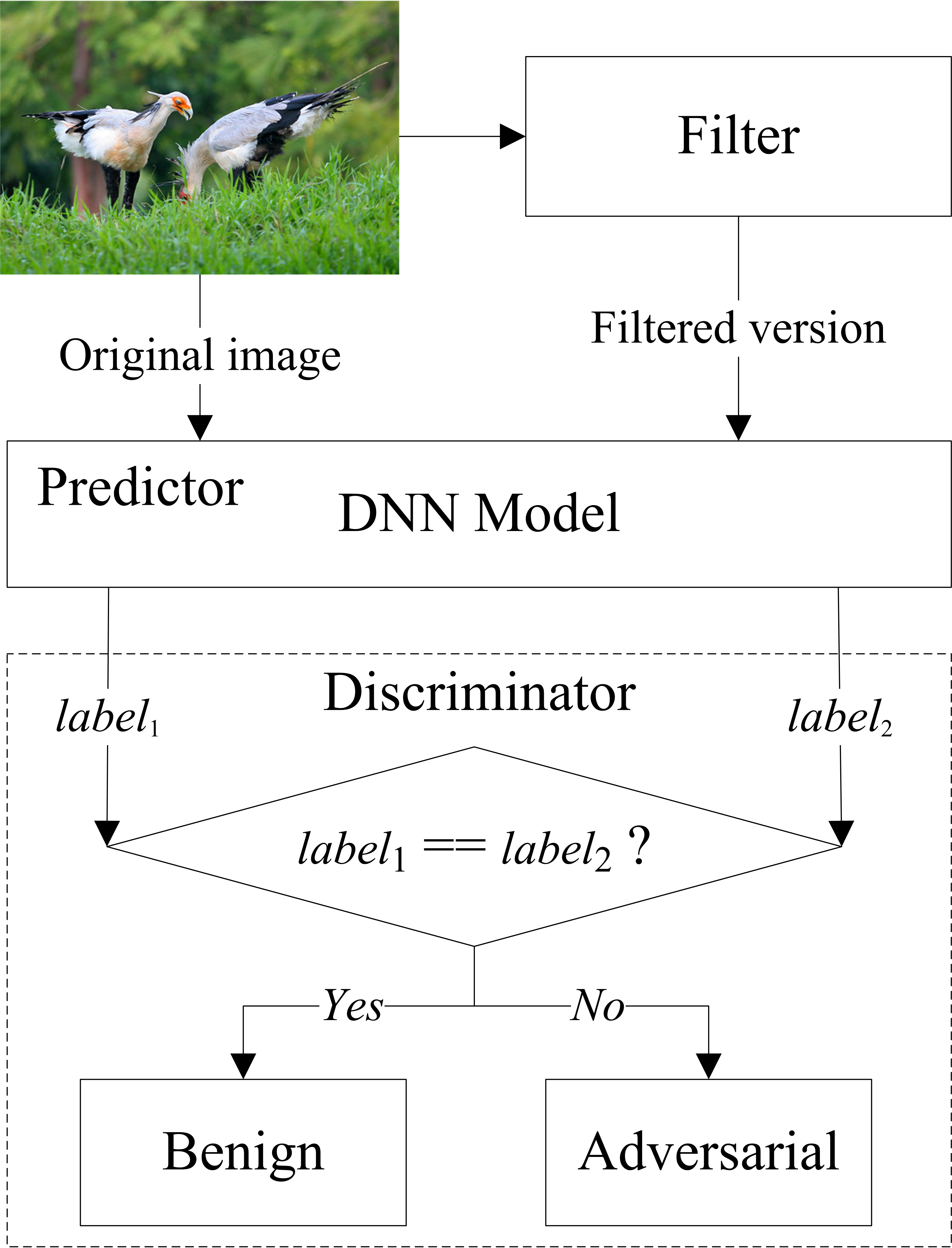}
\caption{Feature-filter framework for adversarial detection.}
\label{fig1}
\end{figure}

Since adversarial examples generated by state-of-the-art attack approaches have rich recessive features to mislead DNN models, they would be much more sensitive to filter out these recessive features than benign examples. A natural detection strategy of adversarial examples is driven by comparing the model's predictions on the original input and its filtered version, as depicted in Fig.~\ref{fig1}. Where, the filter filters out recessive features of the original image as a filtered image. If the original image and its filtered version are classified by the DNN model into different categories, this image would be judged as an adversarial example.

\begin{figure}[h]
\centering

  \begin{subfigure}{0.45\linewidth}
    \includegraphics[width=0.95\linewidth]{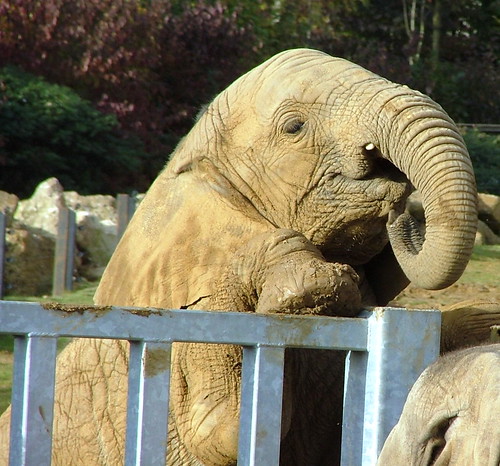}
    \caption{Original image}
  \end{subfigure}
  \begin{subfigure}{0.45\linewidth}
    \includegraphics[width=0.95\linewidth]{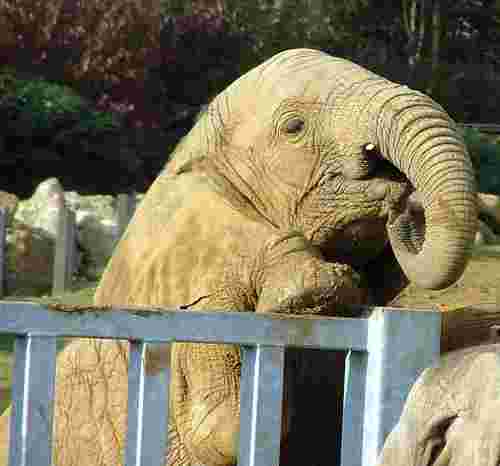}
    \caption{Compressed image}
  \end{subfigure}

\caption{JPEG compression.}
\label{fig2}
\end{figure}

A challenging task is how to reliably filter out the recessive features from the original image. A key idea is inspired by classical JPEG compression based on discrete cosine transform (DCT) \cite{Yahya, Dziugaite, Das}. As depicted in Fig.~\ref{fig2}, the compressed image is still clearly recognizable to humans, even though the feature space is only 1/10 the size of the original image. It indicates that a large number of dominant features are retained, and the contents filtered out almost exclusively include recessive features. Therefore, DCT could be employed as the filter in Fig.~\ref{fig1} to filter out recessive features from input features.

In summary, this paper makes the following contributions:
\begin{itemize}
\item We analyze the reason for the existence of imperceptible adversarial examples, and reveal that the imperceptible adversarial example is the product of abundant recessive features misleading neural networks.

\item To clarify the above explanation, we propose a label-only adversarial detection approach that is referred as feature-filter. Feature-filter can identify adversarial examples by only comparing prediction labels of the DNN on the original image and its filtered version.

\end{itemize}

The remainder of this paper is organized as follows. Section~\ref{sec:relate} introduces the related work about adversarial attacks, defenses and detections. Section~\ref{sec:prel} is the preliminaries about the proposed detection. In Sec.~\ref{sec:method}, we give details about the design and the implementation of the feature-filter. Section~\ref{sec:evalu} evaluates the performance of the DCT-based filter and our detector. Finally, we conclude the paper in Sec.~\ref{sec:conclusion}.

\section{Related work}
\label{sec:relate}
In this section, we briefly review existing works on adversarial attacks, defenses and detections.

\textbf{Adversarial attacks:} The adversarial example is first discovered in exploring properties of neural networks. Szegedy et al. \cite{Szegedy} found that the DNN would misclassify a well-designed image, in which certain perturbations were hardly perceptible to the human eyes. They designed a box-constrained L-BFGS to compute an approximate minimum perturbation matrix.

Many improved works have been subsequently proposed by modifying the adversarial objective function. For example, the fast gradient sign method (FGSM) \cite{Goodfellow} updates each pixel along the direction of the sign of gradient and performs a fast one-step attack. The fast gradient value method (FGVM) \cite{Rozsa} replaces the sign of the gradient with the raw gradient, and generates adversarial examples with a larger local difference. Despite their efficiency, these methods provides only a coarse approximation of the optimal perturbations, resulting in unnecessarily large artifacts.

To improve the imperceptibility of perturbations, DeepFool \cite{Dezfooli} designed an accurate method by moving the input image across its closest decision boundary. Adversarial examples generated by DeepFool have less perturbations in the input image than those generated by L-BFGS and FGSM. Inspired by DeepFool, an universal adversarial attack \cite{Universal} was proposed to reveal important geometric correlations among the high-dimensional decision boundary of DNNs, which proved the transferability of adversarial examples across neural networks.

Among existing adversarial attacks, C\&W \cite{Carlini} attack performs high resilience against the most defense and detection methods. Especially, $L_2$ attack among them is argued to be the most effective to date in white-box threat model. Therefore, most adversarial detectors are evaluated by testing in adversarial examples generated by C\&W attack.

\textbf{Adversarial defenses:} A wide variety of proactive defense approaches have been proposed to harden the neural network. Based on a comprehensive summary, adversarial defenses are grouped into two broad categories: adversarial training and gradient masking \cite{PapernotSok}.

Adversarial training extends the training dataset by adversarial examples with the ground-truth labels and retrains the neural network using this extended dataset. Goodfellow et al. \cite{Goodfellow} claimed that adversarial training could provide an additional regularization benefit and increase the robustness of neural networks. Although this type of defense works against attacks on which the model is trained, it faces potential risks for new attacks.

Gradient masking is an effective defensive strategy against gradient-based adversarial attacks. These attack methods compute first-order derivatives to estimate the sensitivity of the model with respect to inputs. Papernot et al. \cite{PapernotDist} introduced ``defensive distillation" to defend DNNs against adversarial examples. To hide the model's gradients, network distillation replaces its softmax layer with a ``softer" softmax function. In fact, gradients are not essential knowledge for attackers, who easily evade gradient masking using a transferability-based black-box attack \cite{Universal}.

\textbf{Adversarial detections:} Due to the difficulty of adversarial defenses, recent works focus on reactive adversarial detections, which distinguish adversarial images from benign images. Adversarial detections could be grouped into three categories: statistical measurement \cite{Cohen, YangP, FeinmanR}, secondary classifier \cite{MaS, Sperl} and input transformation \cite{XuW, TianS, Yiannis, Bahat}.

Statistical measurement aims to find the statistical differences in the high-dimensional features extracted from DNN layers. NNIF \cite{Cohen} utilized influence functions and nearest neighbors to construct k-NN features, which are employed to train a logistic regression for adversarial detection. ML-LOO \cite{YangP} computes the contribution of features in the prediction of model to construct feature attribution map, and then measures 3 metrics (standard deviation, median absolute deviation and interquartile range) of statistical dispersion in feature attribution map. The experiments revealed a significant difference in the distributions of 3 metrics between benign and adversarial images. Due to the intrinsically unperceptive nature of adversarial examples, statistical measurement seems unlikely to be effective to detect adversarial examples.

Adversarial detections is essentially binary classification problem. A nature idea is to train secondary classifiers to detect if the input is benign or adversarial. NIC \cite{MaS} trains a model for each layer in the DNN to describe the distributions of the provenance and the activation value. By combining the results of all models, NIC makes a joint prediction to determine whether the input image is adversarial or not. In fact, the secondary classifiers also could be fooled, thus, they have the potential vulnerability to second-round adversarial attacks.

The basic idea of input transformation is to measure the disagreement of the target model in predicting the input and its transformed versions. Xu et al. \cite{XuW} proposed a detection strategy, feature squeezing, to squeeze feature space available to attackers. However, they did not specify which features should be squeezed. Kantaros et al. \cite{Yiannis} observed that adversarial examples are sensitive to lossy compression transformations. They proposed VisionGuard to real-time detect large-scale adversarial examples. VisionGuard measured similarity of softmax outputs between the test image and its compressed version using the K-L
divergence measure. When the the K-L divergence is greater than a threshold, the test image is considered an adversarial input. VisionGuard determined the detection threshold based on the training dataset and its adversarial dataset, which makes it inapplicable to data with different distributions. Tian et al. \cite{TianS} exploited a set of rotation operations to yield several transformed versions and analyzed the classification results of these transformed images to determine if the test image is adversarial. Since input transformation can not change neural networks, it is inexpensive and complementary to other defenses and detections.

\section{Preliminaries}
\label{sec:prel}
\subsection{Neural network}
A neural network \cite{HeK, HuangG, Karen, HuJ, Inception} has a hierarchical structure, in which each layer is composed of a set of perceptrons. Perceptrons map inputs to output values with a nonlinear activation function, e.g. sigmoid, tanh, and ReLU. The function of a neural network is formed in a chain.
\begin{equation}
f(x,\theta)=f^{(k)}(\dots f^{(2)}(f^{(1)}(x,\theta_1),\theta_2),\theta_k)
\label{eq1}
\end{equation}
where \emph{x} is the input example and $\theta_i$ is the weight of the $i^\prime$th layer, $i = 1, 2, ... , k$. $f^{(i)}(x,\theta_i)$ is the function of the $i^\prime$th layer of the network.

Convolutional neural network is one of the most widely used neural networks, which consists of alternating convolutional layers and pooling layers. Convolutional neural network has performed incredible successes in image recognition. As a representative network, Inception V3 \cite{Inception} has 23,851,784 parameters and performs 93.70\% top-5 accuracy for the ImageNet datase.

\subsection{Adversarial attack}
Adversarial attack \cite{Goodfellow, YuanX, Dezfooli, Universal, DongJ, Bhattad, SuJ, Laidlaw} is a unique form of attack in deep learning, in which attackers are able to change DNNs' outputs by adding imperceptible perturbations to inputs. Adversarial attack would be formulated as an optimization problem with imperceptibility as the main constraint, that is,
\begin{equation}
\begin{aligned}
{\rm min} \quad  &\lVert \delta \rVert _p \\
s.t. \quad & f(x)=l \\
&f(x')=l' \\
&l \neq l' \\
&x'=x +\delta \in D\\
\end{aligned}
\label{eq2}
\end{equation}
where $\delta$ denotes the perturbation matrix that is added to the benign example \emph{x} for synthesizing the adversarial example $x^\prime$, which remains in the benign domain $D$, and a trained DNN model \emph{f} predicts \emph{x} and $x^\prime$ into different labels ($l\neq l^\prime$), $\rVert \cdot \rVert$ denotes the distance between two data sample.

Among the state-of-the-art attacks, C\&W \cite{Carlini} attacks achieve the best attacking performance comparing with existing-attack algorithms. This set of attacks provide 3 metrics to measure its distortion ($L_0$, $L_2$ and $L_{\infty}$ norms), and $L_2$ norm is the most powerful attack among them. C\&W attacks with $L_2$ norm could be formulated as follows,
\begin{equation}
\begin{aligned}
{\rm min} \quad & \lVert \delta \rVert _2 + c \cdot Loss(x') \\
s.t. \quad & x'=x+\delta \in D\\
\end{aligned}
\label{eq3}
\end{equation}
where $c$ is a hyper-parameter, and \emph{Loss} is defined as
\begin{equation}
Loss(x')={\rm max}({\rm max}\{Z(x')_i : i \neq t\} - Z(x')_t, -k)
\label{eq4}
\end{equation}
where $Z(x)$ is the output of the network before the softmax layer, and a hyper-parameter $k$ encourages the attack to search for an adversarial example $x^\prime$ that will be classified as label $t$ with high confidence.

\subsection{Discrete cosine transform}
Discrete cosine transform (DCT) \cite{Yahya, Dziugaite, Das} is a Fourier-like transform, which performs the cosine-series expansion. Due to its high ``energy compaction'' property, the transformed signal can be easily analyzed using few low-frequency components. Thus, DCT is usually employed to perform decorrelation of the input signal and to present the output in the frequency domain. Among several types of DCT, two-dimensional DCT is the most popular symmetric variation of the transform that operates on images and its inverse. Two-dimensional DCT of an $M \times N$ image $s(x,y)$ are defined as follows.
\begin{equation}
\begin{split}
S(\mu,\nu) =\ &\alpha_\mu \alpha_\nu \sum_{x=0}^{M-1}\sum_{y=0}^{N-1}s(x,y) \\
&\times \cos(\frac{\pi\mu(2x+1)}{2M})\cos(\frac{\pi\nu(2y+1)}{2N})
\end{split}
\label{eq5}
\end{equation}

The inverse of two-dimensional DCT for an $M \times N$ sample:
\begin{equation}
\begin{aligned}
s(x,y)=&\sum_{\mu=0}^{M-1}\sum_{\nu=0}^{N-1}\alpha_\mu\alpha_\nu S(\mu,\nu) \\
&\times \cos(\frac{\pi\mu(2x+1)}{2M})\cos(\frac{\pi\nu(2y+1)}{2N})
\end{aligned}
\label{eq6}
\end{equation}

The matrix $S(\mu,\nu)$ is the DCT coefficients of image $s(x,y)$, whereas, the basis functions are,
\begin{equation}
\begin{split}
&\alpha_\mu=\left\{
\begin{aligned}
1/\sqrt{M}\quad & x=0 \\
\sqrt{2/M}\quad & 0\leq x \leq M-1
\end{aligned}
\right.
\\
&\alpha_\nu=\left\{
\begin{aligned}
1/\sqrt{N}\quad & y=0 \\
\sqrt{2/N}\quad & 0\leq y \leq N-1
\end{aligned}
\right.
\end{split}
\label{eq7}
\end{equation}

The energy of the image concentrates on the low-frequency DCT components. Human eyes have a poor perception in high-frequency DCT domain. Based on this property, Two-dimensional DCT is widely used in lossy compression \cite{Yiannis} and image steganography \cite{DaiH}. In this paper, we apply two-dimensional DCT to availably filter out recessive features from the image.

\section{Methodology}
\label{sec:method}
In this section, we introduce the feature-filter framework and how to design the DCT-based filter to reliably filter out recessive features.
\subsection{Feature-filter framework}
In order to achieve high imperceptibility, state-of-the-art adversarial attacks attempt to enrich recessive features by adding perturbations into original images, which results in a very significant difference in recessive features between benign images and adversarial images. Therefore, adversarial examples is sensitive to the operation of filtering out recessive features, whereas benign examples are not. That is, if the test image is an adversarial example, the DNN-based classifier would predict it and its filtered version into different categories. This idea drove us to design an adversarial example detector by comparing DNN's prediction labels.

Feature-filter employs well-established metamorphic testing to expose adversarial images that lead to inconsistent predictions of the DNN model. As depicted in Fig.~\ref{fig1}, the feature-filter consists of a filter, a predictor, and a discriminator, whose functions are shown below.

(1) Filter: Filter out recessive features of the input image and obtain its filtered version;

(2) Predictor: Enter the test image and its filtered version into DNN model $\emph{f}$ and obtain the DNN's prediction labels;

(3) Discriminator: If the test image and its filtered version are predicted into the same category, the test image is judged to be benign; otherwise, it is judged to be adversarial.

In feature-filter framework, the performance of the proposed detector depends on DNN's prediction on the filtered image. Therefore, a key problem is how to construct a filter so that it can reliably filter out recessive features from the test image. In the next subsection, the construction of the filter is introduced in detail.



\subsection{DCT-based filter}
\begin{figure*}[t]
\centering
\includegraphics[height=1.3in]{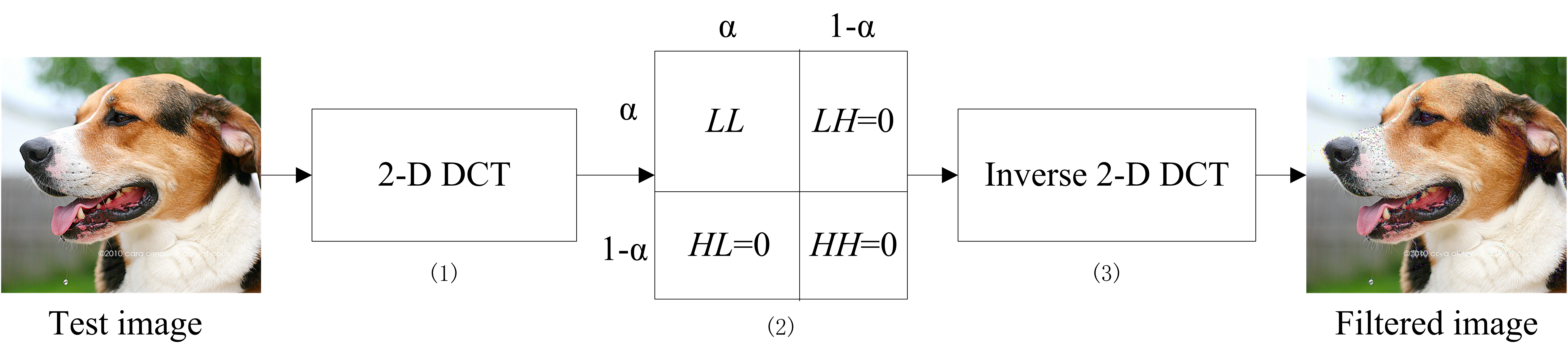}
\caption{Overview of the DCT-based filter.}
\label{fig3}
\end{figure*}

This section describes how we design a DCT-based filter to filter out the recessive features of the test image. As shown in Fig.~\ref{fig2}, the recessive features are usually reserved for high-frequency areas of the image, while the dominant features are reserved for low-frequency areas. Through DCT transform-domain method, we transform spatial-domain image pixels into frequency-domain coefficients and separate the dominant features from the recessive features. By discarding coefficients of high-frequency areas, the filter could filters out most of recessive features. A DCT-based filter is illustrated in Fig.~\ref{fig3}. The details of the filter are listed below. 

(1) Executing two-dimensional DCT, the $M \times N$ image $x$ is transformed into an $M \times N$ matrix \emph{xf} of frequency-domain coefficients;

(2) Generating a dominant feature matrix \emph{lf} by reserving $(\alpha M \times \alpha N)$ low-frequency coefficients and setting all of high-frequency coefficients to 0 in matrix \emph{xf}, where feature reservation ratio $\alpha$ ranges from 0 to 1;

(3) Executing the inverse of two-dimensional DCT, the matrix $lf$ is transformed into a filtered image \emph{xm}, which reserves most of the dominant features and removes a great number of the recessive features.








\section{Experimental evaluation}
\label{sec:evalu}
\subsection{Performance of DCT-based filter}
The design of the feature-filter roots from an explanation for the existence of adversarial examples with imperceptible perturbations, that is, recessive features are involved in neural network decisions. It is critical to reliably filter out recessive features of the test image. In our work, the dominant features and the recessive features are defined from a human-centric perspective. Therefore, the feature-filter should generate filtered versions which are still clearly recognizable to humans.

\begin{figure*}
\centering

  \begin{subfigure}{0.15\linewidth}
    Original images
  \end{subfigure}
  \begin{subfigure}{0.15\linewidth}
    \includegraphics[width=.9\linewidth]{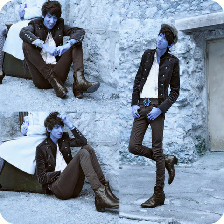}
  \end{subfigure}
  \begin{subfigure}{0.15\linewidth}
    \includegraphics[width=.9\linewidth]{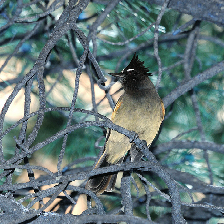}
  \end{subfigure}
  \begin{subfigure}{0.15\linewidth}
    \includegraphics[width=.9\linewidth]{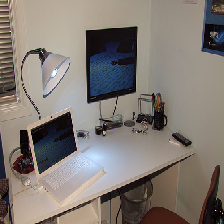}
  \end{subfigure}
  \begin{subfigure}{0.15\linewidth}
    \includegraphics[width=.9\linewidth]{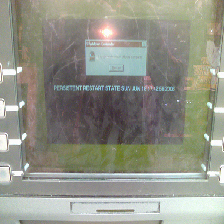}
  \end{subfigure}
  \begin{subfigure}{0.15\linewidth}
    \includegraphics[width=.9\linewidth]{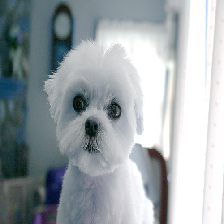}
  \end{subfigure}

  \begin{subfigure}{0.15\linewidth}
    Filtered images
  \end{subfigure}
  \begin{subfigure}{0.15\linewidth}
    \includegraphics[width=.9\linewidth]{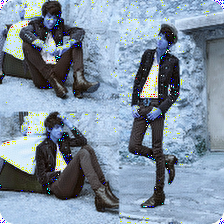}
  \end{subfigure}
  \begin{subfigure}{0.15\linewidth}
    \includegraphics[width=.9\linewidth]{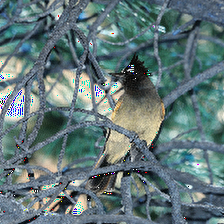}
  \end{subfigure}
  \begin{subfigure}{0.15\linewidth}
    \includegraphics[width=.9\linewidth]{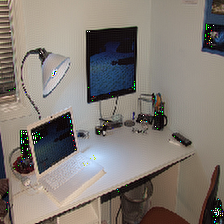}
  \end{subfigure}
  \begin{subfigure}{0.15\linewidth}
    \includegraphics[width=.9\linewidth]{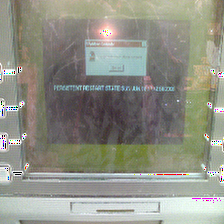}
  \end{subfigure}
  \begin{subfigure}{0.15\linewidth}
    \includegraphics[width=.9\linewidth]{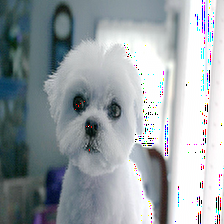}
  \end{subfigure}

\caption{Performance of the DCT-based filter.}

\label{fig4}
\end{figure*}

To test the DCT-based filter, we set feature reservation ratio $\alpha = 0.5$, which means 3/4 of the frequency features are filtered out by the DCT-based filter. As shown in Fig.~\ref{fig4}, five random test images are entered into the DCT-based filter and we obtain their filtered versions. Even if most of the high-frequency features are moved by the DCT-based filter, human eyes still clearly identify test images and filtered versions as the same category. It indicates that the moved features are not perceived by the human eyes.

Whether the DCT-based filter can reliably filter out recessive features from the test image? To verify the performance of the DCT-based filter, we conducted an extensive user study based on human eye evaluation. In the test, the DCT-based filter generates 300 filtered images at multiple feature reservation ratios, which are from 30 benign images in 10 categories in ImageNet dataset \cite{imagenet}. We recruit a total of 50 participants to label these filtered versions. For each trail, filtered images are shown on the screen at a fixed size and the order of images is random. In ImageNet dataset, each image contains more than one object, and only one object category is labeled in each image. In order to clarify ambiguity in the evaluation, we design the following test.

\textbf{Test:} Each participant is required to label 300 filtered images generated by the DCT-based filter. The participants are allow to identify multiple (up to 5) objects in an image. As long as one of the objects indeed corresponded to the ground-truth label, this filtered image would be judged to reserve the enough dominant features. We calculate the proportion of trails where filtered images are labeled as a ground-truth label. Higher proportion denotes better performance of the DCT-based filter.

\textbf{Results:} Table \ref{table_1} lists the proportion of filtered images being labeled correctly at 4 feature reservation ratios $\alpha$. Most of the filtered images are still recognizable to human eyes, which indicates human eyes are not sensitive to the high-frequency features of images. In particular, when $\alpha$ = 0.6, i.e. 64\% of the high-frequency features are filtered out, the filtered images are still correctly identified 96.53\% of the time. That is to say, human eyes rarely use high-frequency features to understand images. We also observe that the proportion slowly increases as $\alpha$ increases, which denotes the high-frequency features only contain a small number of comprehensible features to humans (dominant features). The results show that the DCT-based filter can reliably filter out recessive features from the test image.

\begin{table}
  \centering
  \begin{tabular}{@{}lccccc@{}}
    \toprule
    $\alpha$ & 0.9 & 0.8 & 0.7 & 0.6 \\
    \midrule
    Proportion & \textbf{98.73\%} & 98.33\% & 97.67\% & 96.53\% \\
    \bottomrule
  \end{tabular}
  \caption{Proportion of filtered images that are correctly labeled by human eyes.}
  \label{table_1}
\end{table}

\subsection{Detection against C\&W attack}
The design of the feature-filter is based on this assumption that the attacker wants adversarial examples with imperceptible perturbations. To test its detection performance, we need to generate a set of adversarial examples with very imperceptible perturbations. Among state-of-the-art adversarial attacks, C\&W attack is proven to be a more powerful attack, in which $L_2$ attack can generate higher quality adversarial examples. C\&W attack can bypass multiple types of detectors. Similar results are reported in the literature \cite{CWb} where C\&W attack bypassed all of the ten detection methods. Therefore, we focus on our experiments with the challenging C\&W attack ($L_2$ norm) on the CIFAR-10 over Carlini network \cite{Carlini} and ImageNet over Inception V3.

We evaluate the feature-filter from true-positive rate (TPR) and true-negative rate (TNR). TPR measures the proportion of adversarial examples that are correctly identified by the detector, while TNR measures the proportion of benign examples that are correctly identified by the detector. The experiment randomly select 500 CIFAR-10 adversarial examples and 50 ImageNet adversarial examples generated by C\&W attack to compute TPR and TNR. Notes that the results exclude the failed adversarial examples and over-fitting examples from consideration.

Table \ref{table_2} lists the results of TPR and TNR at multiple feature reservation ratios $\alpha$. In the best case, the detection rate of adversarial examples is as high as 98.20\% for CIFAR-10 when $\alpha$ = 0.8 and 100.00\% for ImageNet when $\alpha$ = 0.7. In addition, we also observe that TNR decreases as $\alpha$ decreases, which is consistent with the facts that more feature loss reduces the accuracy of neural networks.

\begin{table}[ht]
\centering
\begin{tabular}{@{}ccccc@{}}
\toprule
\multirow{2}*{$\alpha$}   & \multicolumn{2}{c}{CIFAR-10} & \multicolumn{2}{c}{ImageNet} \\ \cline{2-5}
                          & TPR & TNR                     & TPR & TNR \\ \midrule
0.95                      & 97.00\% & \textbf{98.80\%}    & 72.00\% & \textbf{100.00\%} \\
0.90                      & \textbf{98.20}\% & 97.00\%    & 84.00\% & 100.00\% \\
0.85                      & 97.80\% & 96.80\%             & 96.00\% & 100.00\% \\
0.80                      & 98.20\% & 94.20\%             & 98.00\% & 100.00\% \\
0.75                      & 98.00\% & 92.60\%             & 98.00\% & 100.00\% \\
0.70                      & 98.20\% & 89.20\%             & \textbf{100.00\%} & 98.00\% \\
0.65                      & 97.00\% & 83.00\%             & 100.00\% & 98.00\% \\
0.60                      & 95.60\% & 79.20\%             & 100.00\% & 90.00\% \\
0.55                      & 96.60\% & 69.60\%             & 100.00\% & 92.00\% \\
0.50                      & 96.20\% & 66.00\%             & 100.00\% & 90.00\% \\
\bottomrule
\end{tabular}
\caption{TPR and TNR at multiple feature reservation ratios $\alpha$.}
\label{table_2}
\end{table}

We compare the feature-filter with previous adversarial detection approaches based on image transformation, e.g., bit depth reduction\cite{XuW}, spatial smoothing (local smoothing and non-local smoothing)\cite{XuW}, and rotation \cite{TianS}. Table \ref{table_3} lists the results of TPR and TNR for several detectors built upon single transformation on the CIFAR-10 dataset. The results show that DCT transform is a more effective approach than other image transformation to distinguish adversarial examples from benign examples.
\begin{table}[htb]
\centering
\begin{tabular}{@{}cccc@{}}
\toprule
Approaches                          & Parameters    & TPR               & TNR               \\ \midrule
\multirow{5}*{Feature-Filter}       & 0.90          & \textbf{98.20\%}  & \textbf{97.00\%}  \\ 
                                    & 0.80          & 98.20\%           & 94.20\%           \\ 
                                    & 0.70          & 98.20\%           & 89.20\%           \\ 
                                    & 0.60          & 95.60\%           & 79.20\%           \\ 
                                    & 0.50          & 96.20\%           & 66.00\%           \\ \midrule
\multirow{5}*{Bit Depth Reduction}  & 1-bit         & 89.15\%           & 45.11\%           \\ 
                                    & 2-bit         & 92.22\%           & 78.31\%           \\ 
                                    & 3-bit         & \textbf{93.88\%}  & 92.69\%           \\ 
                                    & 4-bit         & 89.00\%           & 97.52\%           \\ 
                                    & 5-bit         & 85.75\%           & \textbf{98.76\%}  \\ \midrule
\multirow{3}*{Median Smoothing}     & 2$\times$2    & \textbf{95.09\%}  & \textbf{82.03\%}  \\ 
                                    & 3$\times$3    & 94.07\%           & 66.17\%           \\ 
                                    & 4$\times$4    & 89.80\%           & 43.87\%           \\ \midrule
\multirow{4}*{Non-local Mean}       & 11-3-2        & 86.30\%           & 99.25\%           \\ 
                                    & 11-3-4        & \textbf{92.73\%}  & 96.41\%           \\ 
                                    & 13-3-2        & 89.72\%           & \textbf{99.26\%}  \\ 
                                    & 13-3-4        & 90.63\%           & 97.03\%           \\ \midrule
\multirow{4}*{Rotation}             & -20           & 88.48\%           & 60.22\%           \\ 
                                    & -10           & 91.60\%           & 84.01\%           \\ 
                                    & 10            & \textbf{93.33\%}  & \textbf{85.50\%}  \\ 
                                    & 20            & 91.17\%           & 65.30\%           \\
\bottomrule
\end{tabular}
\caption{Comparison of TPR and TNR based on several image transformation on CIFAR-10.}
\label{table_3}
\end{table}

\begin{figure}[htb]
\centering
\includegraphics[width=0.8\columnwidth]{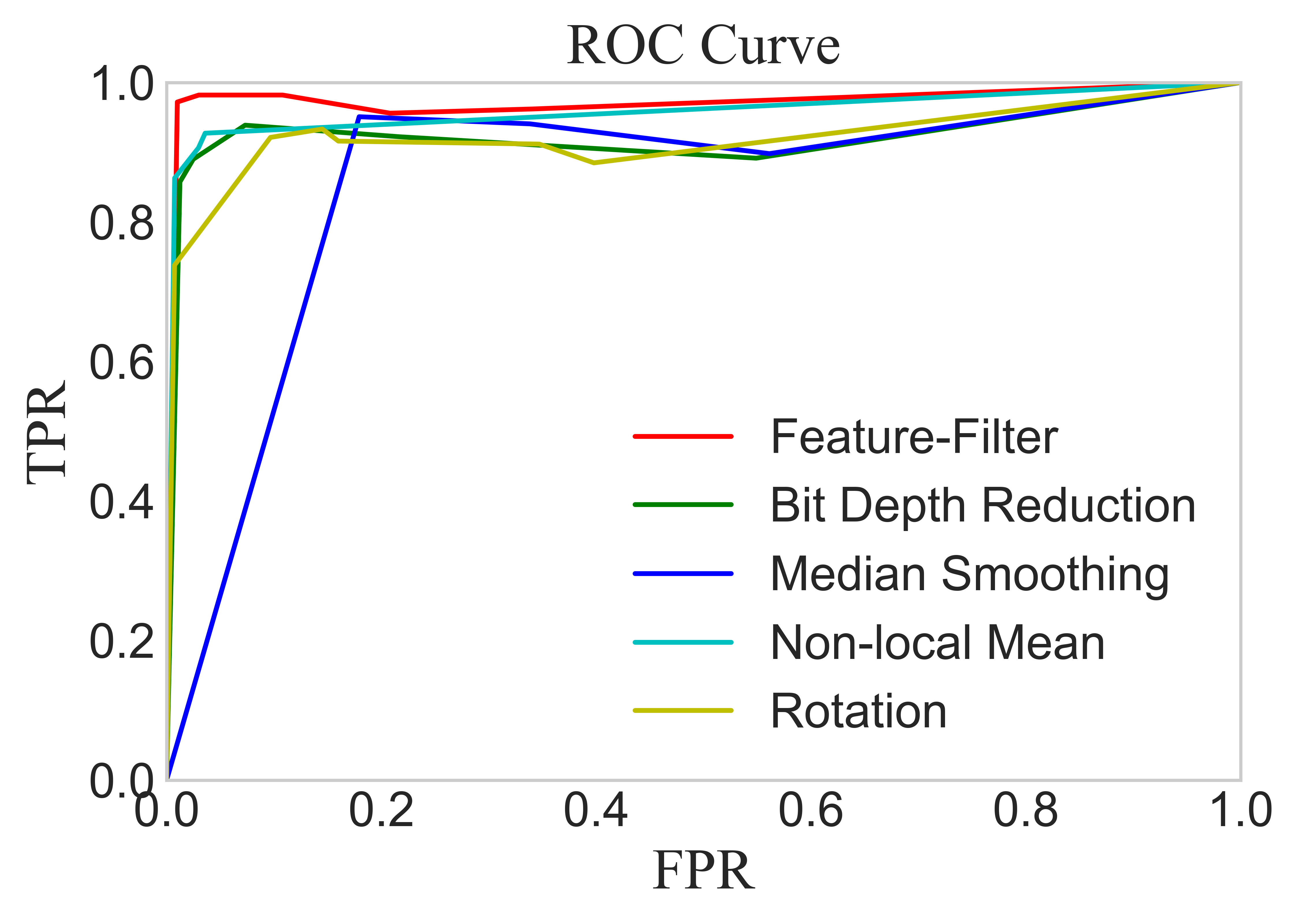}
\caption{Comparison of ROC curves.}
\label{fig5}
\end{figure}

We employ the Receiver Operating Characteristic (ROC) curve and the corresponding Area Under Curve (AUC) to evaluate the detection performance on the CIFAR-10 dataset. Figure~\ref{fig5} plots the ROC curves for several detectors on the task of detecting adversarial images. Table~\ref{table_4} lists the AUC values for several detectors based on image transformation.

The high detection rate shows that filtering out recessive features could significantly change DNN's prediction labels of imperceptible adversarial examples. The results suggest that recessive features of misleading DNNs result in imperceptible adversarial examples, and the adversarial attack enriches these recessive features.

\begin{table}[htb]
\centering
\begin{tabular}{@{}cc@{}}
\toprule
Approaches & AUC \\
\midrule
Feature-Filter & \textbf{96.23\%} \\
Bit Depth Reduction & 92.18\% \\
Median Smoothing & 85.70\% \\
Non-local Mean & 95.76\% \\
Rotation & 91.95\% \\
\bottomrule
\end{tabular}
\caption{Comparison of AUC values.}
\label{table_4}
\end{table}

\begin{table*}[ht]
\centering
\begin{tabular}{@{}ccccccc@{}}
\toprule
\multirow{2}{*}{$\alpha$} & \multicolumn{2}{c}{Gaussian} & \multicolumn{2}{c}{Poisson} & \multicolumn{2}{c}{Salt\&Pepper}\\
\cline{2-7}
 & Top-1 & Top-5 & Top-1 & Top-5 & Top-1 & Top-5\\
\midrule
0.95 & \textbf{79.42}\% & \textbf{93.31\%} & \textbf{80.87\%} & \textbf{94.68\%} & \textbf{79.29\%} & \textbf{93.62\%}\\
0.90 & 79.19\% & 93.20\% & 79.76\% & 94.59\% & 75.61\% & 93.01\%\\
0.85 & 78.08\% & 92.93\% & 79.50\% & 94.18\% & 73.98\% & 90.94\%\\
0.80 & 79.38\% & 92.20\% & 79.42\% & 93.37\% & 74.64\% & 89.77\%\\
0.75 & 79.13\% & 92.47\% & 79.29\% & 94.05\% & 72.32\% & 88.67\%\\
0.70 & 77.72\% & 91.53\% & 78.38\% & 93.21\% & 71.78\% & 87.92\%\\
0.65 & 77.51\% & 91.39\% & 77.52\% & 92.45\% & 71.29\% & 88.24\%\\
0.60 & 74.60\% & 91.23\% & 75.90\% & 91.05\% & 71.33\% & 88.79\%\\
0.55 & 73.08\% & 89.76\% & 75.67\% & 90.85\% & 70.05\% & 88.28\%\\
0.50 & 70.79\% & 88.62\% & 72.53\% & 89.66\% & 67.62\% & 86.19\%\\
\bottomrule
\end{tabular}
\caption{Natural noise detection on the ImageNet.}
\label{table_5}
\end{table*}

\subsection{Natural noise detection}
Adversarial examples are a kind of images with special noises, while there are a large number of images with natural noise in reality. A good adversarial detector should be able to tolerate natural noises, which would identify natural noise examples to be benign rather than adversarial.

The section verifies that our detector can correctly identify natural noise images as benign ones. We add 3 types of noises, e.g., Gaussian, Poisson, and salt\&pepper to generate natural noise images. The natural noise detection performance is evaluated on a pre-trained Inception V3 \cite{Inception} model in the ImageNet classification task. Notes that natural noises here refer to the perturbations which do not change the prediction of the neural network, so that distinguishing them from adversarial perturbations.

Table \ref{table_5} lists the accuracy of the feature-filter in detecting 3 types of natural noise images. Due to multiple objects in an image of the ImageNet dataset, We test top-1 and top-5 to evaluate our detector performance on natural noise images. Higher accuracy denotes better tolerance of the detector to the natural noise. As shown in Tab.~\ref{table_5}, the top-5 accuracy is significantly higher than that of the top-1.  Particularly, the top-5 accuracy of the feature-filter is close to 95\% for Poisson noise. And our detector shows an approximate performance for 3 types of natural noises, which can stably avoid the interference of natural noise. We also see that the accuracy of the feature-filter slowly decreases as $\alpha$ decreases. The phenomenon reveals that high-frequency features only contain a small amount of dominant features. Therefore, the DCT-based filter could be regarded as a reliable tool to filter out the recessive features from the test image.

\subsection{Computation cost}
Apart from detection rate, efficiency is also an important parameter for a good adversarial detector. The feature-filter consists of a filter, a predictor and a discriminator. The filter is based on two-dimensional DCT to generate the filtered image of the test image. As defined in Eqs.~\ref{eq5} and \ref{eq6}, the time complexity of DCT and its inverse transform is $O(M \times N)$ for an $M \times N$ image. The predictor is made up of the target DNN, which performs two predictions on the test image and its filtered version. Thus, computation cost of the predictor depends on the efficiency of the neural network. The discriminator only gives the decision according to two predicted labels by the predictor, thus its time complexity is $O(1)$.

The actual running time depends on many factors including experimental setup, programming skill, etc. Our test is implemented on an Intel Core I5 CPU 2.30 GHz, NVIDIA GeForce GTX 1060 and 8.0 GB of RAM computer that run on Windows 10 and Spyder (Python 3.6). We choose 1,000 random images to test the running time of the feature-filter. The average running time is 0.0351 second for a $224 \times 224 \times 3$ ImageNet image and 0.0028 second for a $32 \times 32 \times 3$ CIFAR-10 image. Due to the low computation cost, the feature-filter is considered a real-time adversarial detector.

\section{Conclusion}
\label{sec:conclusion}
The explanation of adversarial examples remains an open issue. As a human-centric phenomenon, we divide the image feature into dominant features and recessive features. We reveal that the existence of imperceptible adversarial examples is due to the DNN's use of the recessive features that mislead neural networks. Adversarial attacks attempt to enrich these recessive features so that the perturbed example can ``fool'' the DNN without noticeable artifacts.

According to above explanation, adversarial examples should be more sensitive to the operation of filtering out recessive features than benign/noisy examples. Inspired by this idea, we propose an adversarial detection approach named feature-filter. The feature-filter determines whether the test image is adversarial or benign by comparing the DNN’s prediction labels on the test input and its filtered version. To filter out recessive features, we introduce two-dimensional DCT to separate the dominant features and the recessive features. The experimental results show that the DCT-based filter can reliably filter out the recessive features of the image, and the feature-filter is a label-only tool to detect adversarial examples in real time.

The effectiveness of the feature-filter stems from a reliable filtering of recessive features. As discussed in Sec.~\ref{sec:evalu}, two-dimensional DCT is an approximate method to separate the recessive features from the dominant features. Future works are to discover more reliable feature filtering methods or combine the feature-filter with other methods, so that the adversarial detector is more accurate and robust.

{\small
\bibliographystyle{ieee_fullname}
\bibliography{PaperForReview}

\begin{thebibliography}{10}\itemsep=-1pt

\bibitem{Bahat}
Yuval Bahat, Michal Irani, and Gregory Shakhnarovich.
\newblock Natural and adversarial error detection using invariance to image
  transformations.
\newblock {\em CoRR}, abs/1902.00236, 2019.

\bibitem{Bhattad}
Anand Bhattad, Min~Jin Chong, Kaizhao Liang, Bo Li, and David~A. Forsyth.
\newblock Unrestricted adversarial examples via semantic manipulation.
\newblock In {\em 8th International Conference on Learning Representations,
  {ICLR} 2020, Addis Ababa, Ethiopia, April 26-30, 2020}. OpenReview.net, 2020.

\bibitem{BlancoG}
Ramon Blanco{-}Gonzalo, Oscar Miguel{-}Hurtado, Chiara Lunerti, Richard~M.
  Guest, Barbara Corsetti, Elakkiya Ellavarason, and Raul
  S{\'{a}}nchez{-}Reillo.
\newblock Biometric systems interaction assessment: The state of the art.
\newblock {\em {IEEE} Trans. Hum. Mach. Syst.}, 49(5):397--410, 2019.

\bibitem{Bubeck}
S{\'{e}}bastien Bubeck, Eric Price, and Ilya~P. Razenshteyn.
\newblock Adversarial examples from computational constraints.
\newblock {\em CoRR}, abs/1805.10204, 2018.

\bibitem{CWb}
Nicholas Carlini and David~A. Wagner.
\newblock Adversarial examples are not easily detected: Bypassing ten detection
  methods.
\newblock In Bhavani~M. Thuraisingham, Battista Biggio, David~Mandell Freeman,
  Brad Miller, and Arunesh Sinha, editors, {\em Proceedings of the 10th {ACM}
  Workshop on Artificial Intelligence and Security, AISec@CCS 2017, Dallas, TX,
  USA, November 3, 2017}, pages 3--14. {ACM}, 2017.

\bibitem{Carlini}
Nicholas Carlini and David~A. Wagner.
\newblock Towards evaluating the robustness of neural networks.
\newblock In {\em 2017 {IEEE} Symposium on Security and Privacy, {SP} 2017, San
  Jose, CA, USA, May 22-26, 2017}, pages 39--57. {IEEE} Computer Society, 2017.

\bibitem{Chiaroni}
Florent Chiaroni, Mohamed{-}Cherif Rahal, Nicolas Hueber, and
  Fr{\'{e}}d{\'{e}}ric Dufaux.
\newblock Self-supervised learning for autonomous vehicles perception: {A}
  conciliation between analytical and learning methods.
\newblock {\em {IEEE} Signal Process. Mag.}, 38(1):31--41, 2021.

\bibitem{CohenG}
Gilad Cohen, Guillermo Sapiro, and Raja Giryes.
\newblock Detecting adversarial samples using influence functions and nearest
  neighbors.
\newblock In {\em 2020 {IEEE/CVF} Conference on Computer Vision and Pattern
  Recognition, {CVPR} 2020, Seattle, WA, USA, June 13-19, 2020}, pages
  14441--14450. {IEEE}, 2020.

\bibitem{Cohen}
Gilad Cohen, Guillermo Sapiro, and Raja Giryes.
\newblock Detecting adversarial samples using influence functions and nearest
  neighbors.
\newblock In {\em 2020 {IEEE/CVF} Conference on Computer Vision and Pattern
  Recognition, {CVPR} 2020, Seattle, WA, USA, June 13-19, 2020}, pages
  14441--14450. {IEEE}, 2020.

\bibitem{DaiH}
Hong{-}Zhu Dai, Jie Cheng, and Yafeng Li.
\newblock A novel steganography algorithm based on quantization table
  modification and image scrambling in {DCT} domain.
\newblock {\em Int. J. Pattern Recognit. Artif. Intell.},
  35(1):2154001:1--2154001:18, 2021.

\bibitem{Das}
Nilaksh Das, Madhuri Shanbhogue, Shang{-}Tse Chen, Fred Hohman, Li Chen,
  Michael~E. Kounavis, and Duen~Horng Chau.
\newblock Keeping the bad guys out: Protecting and vaccinating deep learning
  with {JPEG} compression.
\newblock {\em CoRR}, abs/1705.02900, 2017.

\bibitem{Dziugaite}
Gintare~Karolina Dziugaite, Zoubin Ghahramani, and Daniel~M. Roy.
\newblock A study of the effect of {JPG} compression on adversarial images.
\newblock {\em CoRR}, abs/1608.00853, 2016.

\bibitem{FeinmanR}
Reuben Feinman, Ryan~R. Curtin, Saurabh Shintre, and Andrew~B. Gardner.
\newblock Detecting adversarial samples from artifacts.
\newblock {\em CoRR}, abs/1703.00410, 2017.

\bibitem{Goodfellow}
Ian~J. Goodfellow, Jonathon Shlens, and Christian Szegedy.
\newblock Explaining and harnessing adversarial examples.
\newblock In Yoshua Bengio and Yann LeCun, editors, {\em 3rd International
  Conference on Learning Representations, {ICLR} 2015, San Diego, CA, USA, May
  7-9, 2015, Conference Track Proceedings}, 2015.

\bibitem{HeK}
Kaiming He, Xiangyu Zhang, Shaoqing Ren, and Jian Sun.
\newblock Deep residual learning for image recognition.
\newblock In {\em 2016 {IEEE} Conference on Computer Vision and Pattern
  Recognition, {CVPR} 2016, Las Vegas, NV, USA, June 27-30, 2016}, pages
  770--778. {IEEE} Computer Society, 2016.

\bibitem{HuJ}
Jie Hu, Li Shen, and Gang Sun.
\newblock Squeeze-and-excitation networks.
\newblock In {\em 2018 {IEEE} Conference on Computer Vision and Pattern
  Recognition, {CVPR} 2018, Salt Lake City, UT, USA, June 18-22, 2018}, pages
  7132--7141. {IEEE} Computer Society, 2018.

\bibitem{HuangG}
Gao Huang, Zhuang Liu, Laurens van~der Maaten, and Kilian~Q. Weinberger.
\newblock Densely connected convolutional networks.
\newblock In {\em 2017 {IEEE} Conference on Computer Vision and Pattern
  Recognition, {CVPR} 2017, Honolulu, HI, USA, July 21-26, 2017}, pages
  2261--2269. {IEEE} Computer Society, 2017.

\bibitem{Ilyas}
Andrew Ilyas, Shibani Santurkar, Dimitris Tsipras, Logan Engstrom, Brandon
  Tran, and Aleksander Madry.
\newblock Adversarial examples are not bugs, they are features.
\newblock In Hanna~M. Wallach, Hugo Larochelle, Alina Beygelzimer, Florence
  d'Alch{\'{e}}{-}Buc, Emily~B. Fox, and Roman Garnett, editors, {\em Advances
  in Neural Information Processing Systems 32: Annual Conference on Neural
  Information Processing Systems 2019, NeurIPS 2019, December 8-14, 2019,
  Vancouver, BC, Canada}, pages 125--136, 2019.

\bibitem{Yiannis}
Yiannis Kantaros, Taylor~J. Carpenter, Sangdon Park, Radoslav Ivanov, Sooyong
  Jang, Insup Lee, and James Weimer.
\newblock Visionguard: Runtime detection of adversarial inputs to perception
  systems.
\newblock {\em CoRR}, abs/2002.09792, 2020.

\bibitem{Kurakin}
Alexey Kurakin, Ian~J. Goodfellow, and Samy Bengio.
\newblock Adversarial machine learning at scale.
\newblock In {\em 5th International Conference on Learning Representations,
  {ICLR} 2017, Toulon, France, April 24-26, 2017, Conference Track
  Proceedings}. OpenReview.net, 2017.

\bibitem{Laidlaw}
Cassidy Laidlaw, Sahil Singla, and Soheil Feizi.
\newblock Perceptual adversarial robustness: Defense against unseen threat
  models.
\newblock {\em CoRR}, abs/2006.12655, 2020.

\bibitem{DongJ}
Jinfeng Li, Shouling Ji, Tianyu Du, Bo Li, and Ting Wang.
\newblock Textbugger: Generating adversarial text against real-world
  applications.
\newblock In {\em 26th Annual Network and Distributed System Security
  Symposium, {NDSS} 2019, San Diego, California, USA, February 24-27, 2019}.
  The Internet Society, 2019.

\bibitem{Lovisotto}
Giulio Lovisotto, Henry Turner, Ivo Sluganovic, Martin Strohmeier, and Ivan
  Martinovic.
\newblock {SLAP:} improving physical adversarial examples with short-lived
  adversarial perturbations.
\newblock In {\em 30th {USENIX} Security Symposium, {USENIX} Security 2021,
  August 11-13, 2021}, pages 1865--1882, 2021.

\bibitem{MaS}
Shiqing Ma, Yingqi Liu, Guanhong Tao, Wen{-}Chuan Lee, and Xiangyu Zhang.
\newblock {NIC:} detecting adversarial samples with neural network invariant
  checking.
\newblock In {\em 26th Annual Network and Distributed System Security
  Symposium, {NDSS} 2019, San Diego, California, USA, February 24-27, 2019}.
  The Internet Society, 2019.

\bibitem{Mahloujifar}
Saeed Mahloujifar, Dimitrios~I. Diochnos, and Mohammad Mahmoody.
\newblock The curse of concentration in robust learning: Evasion and poisoning
  attacks from concentration of measure.
\newblock In {\em The Thirty-Third {AAAI} Conference on Artificial
  Intelligence, {AAAI} 2019, The Thirty-First Innovative Applications of
  Artificial Intelligence Conference, {IAAI} 2019, The Ninth {AAAI} Symposium
  on Educational Advances in Artificial Intelligence, {EAAI} 2019, Honolulu,
  Hawaii, USA, January 27 - February 1, 2019}, pages 4536--4543. {AAAI} Press,
  2019.

\bibitem{MirskyY}
Yisroel Mirsky, Tomer Doitshman, Yuval Elovici, and Asaf Shabtai.
\newblock Kitsune: An ensemble of autoencoders for online network intrusion
  detection.
\newblock {\em CoRR}, abs/1802.09089, 2018.

\bibitem{Universal}
Seyed{-}Mohsen Moosavi{-}Dezfooli, Alhussein Fawzi, Omar Fawzi, and Pascal
  Frossard.
\newblock Universal adversarial perturbations.
\newblock In {\em 2017 {IEEE} Conference on Computer Vision and Pattern
  Recognition, {CVPR} 2017, Honolulu, HI, USA, July 21-26, 2017}, pages 86--94.
  {IEEE} Computer Society, 2017.

\bibitem{Dezfooli}
Seyed{-}Mohsen Moosavi{-}Dezfooli, Alhussein Fawzi, and Pascal Frossard.
\newblock Deepfool: {A} simple and accurate method to fool deep neural
  networks.
\newblock In {\em 2016 {IEEE} Conference on Computer Vision and Pattern
  Recognition, {CVPR} 2016, Las Vegas, NV, USA, June 27-30, 2016}, pages
  2574--2582. {IEEE} Computer Society, 2016.

\bibitem{PapernotSok}
Nicolas Papernot, Patrick~D. McDaniel, Arunesh Sinha, and Michael~P. Wellman.
\newblock Towards the science of security and privacy in machine learning.
\newblock {\em CoRR}, abs/1611.03814, 2016.

\bibitem{Papernot}
Nicolas Papernot, Patrick~D. McDaniel, Xi Wu, Somesh Jha, and Ananthram Swami.
\newblock Distillation as a defense to adversarial perturbations against deep
  neural networks.
\newblock In {\em {IEEE} Symposium on Security and Privacy, {SP} 2016, San
  Jose, CA, USA, May 22-26, 2016}, pages 582--597. {IEEE} Computer Society,
  2016.

\bibitem{PapernotDist}
Nicolas Papernot, Patrick~D. McDaniel, Xi Wu, Somesh Jha, and Ananthram Swami.
\newblock Distillation as a defense to adversarial perturbations against deep
  neural networks.
\newblock In {\em {IEEE} Symposium on Security and Privacy, {SP} 2016, San
  Jose, CA, USA, May 22-26, 2016}, pages 582--597. {IEEE} Computer Society,
  2016.

\bibitem{PeiK}
Kexin Pei, Yinzhi Cao, Junfeng Yang, and Suman Jana.
\newblock Deepxplore: Automated whitebox testing of deep learning systems.
\newblock In {\em Proceedings of the 26th Symposium on Operating Systems
  Principles, Shanghai, China, October 28-31, 2017}, pages 1--18. {ACM}, 2017.

\bibitem{Rozsa}
Andras Rozsa, Ethan~M. Rudd, and Terrance~E. Boult.
\newblock Adversarial diversity and hard positive generation.
\newblock In {\em 2016 {IEEE} Conference on Computer Vision and Pattern
  Recognition Workshops, {CVPR} Workshops 2016, Las Vegas, NV, USA, June 26 -
  July 1, 2016}, pages 410--417. {IEEE} Computer Society, 2016.

\bibitem{imagenet}
Olga Russakovsky, Jia Deng, Hao Su, Jonathan Krause, Sanjeev Satheesh, Sean Ma,
  Zhiheng Huang, Andrej Karpathy, Aditya Khosla, Michael~S. Bernstein,
  Alexander~C. Berg, and Fei{-}Fei Li.
\newblock Imagenet large scale visual recognition challenge.
\newblock {\em Int. J. Comput. Vis.}, 115(3):211--252, 2015.

\bibitem{Shafahi}
Ali Shafahi, W.~Ronny Huang, Christoph Studer, Soheil Feizi, and Tom Goldstein.
\newblock Are adversarial examples inevitable?
\newblock In {\em 7th International Conference on Learning Representations,
  {ICLR} 2019, New Orleans, LA, USA, May 6-9, 2019}. OpenReview.net, 2019.

\bibitem{Shaham}
Uri Shaham, Yutaro Yamada, and Sahand Negahban.
\newblock Understanding adversarial training: Increasing local stability of
  supervised models through robust optimization.
\newblock {\em Neurocomputing}, 307:195--204, 2018.

\bibitem{Karen}
Karen Simonyan and Andrew Zisserman.
\newblock Very deep convolutional networks for large-scale image recognition.
\newblock In Yoshua Bengio and Yann LeCun, editors, {\em 3rd International
  Conference on Learning Representations, {ICLR} 2015, San Diego, CA, USA, May
  7-9, 2015, Conference Track Proceedings}, 2015.

\bibitem{Sperl}
Philip Sperl, Ching{-}Yu Kao, Peng Chen, Xiao Lei, and Konstantin
  B{\"{o}}ttinger.
\newblock {DLA:} dense-layer-analysis for adversarial example detection.
\newblock In {\em {IEEE} European Symposium on Security and Privacy, EuroS{\&}P
  2020, Genoa, Italy, September 7-11, 2020}, pages 198--215. {IEEE}, 2020.

\bibitem{SuJ}
Jiawei Su, Danilo~Vasconcellos Vargas, and Kouichi Sakurai.
\newblock One pixel attack for fooling deep neural networks.
\newblock {\em {IEEE} Trans. Evol. Comput.}, 23(5):828--841, 2019.

\bibitem{Inception}
Christian Szegedy, Vincent Vanhoucke, Sergey Ioffe, Jonathon Shlens, and
  Zbigniew Wojna.
\newblock Rethinking the inception architecture for computer vision.
\newblock In {\em 2016 {IEEE} Conference on Computer Vision and Pattern
  Recognition, {CVPR} 2016, Las Vegas, NV, USA, June 27-30, 2016}, pages
  2818--2826. {IEEE} Computer Society, 2016.

\bibitem{Szegedy}
Christian Szegedy, Wojciech Zaremba, Ilya Sutskever, Joan Bruna, Dumitru Erhan,
  Ian~J. Goodfellow, and Rob Fergus.
\newblock Intriguing properties of neural networks.
\newblock In Yoshua Bengio and Yann LeCun, editors, {\em 2nd International
  Conference on Learning Representations, {ICLR} 2014, Banff, AB, Canada, April
  14-16, 2014, Conference Track Proceedings}, 2014.

\bibitem{TianS}
Shixin Tian, Guolei Yang, and Ying Cai.
\newblock Detecting adversarial examples through image transformation.
\newblock In Sheila~A. McIlraith and Kilian~Q. Weinberger, editors, {\em
  Proceedings of the Thirty-Second {AAAI} Conference on Artificial
  Intelligence, (AAAI-18), the 30th innovative Applications of Artificial
  Intelligence (IAAI-18), and the 8th {AAAI} Symposium on Educational Advances
  in Artificial Intelligence (EAAI-18), New Orleans, Louisiana, USA, February
  2-7, 2018}, pages 4139--4146. {AAAI} Press, 2018.

\bibitem{Tram}
Florian Tram{\`{e}}r, Alexey Kurakin, Nicolas Papernot, Ian~J. Goodfellow, Dan
  Boneh, and Patrick~D. McDaniel.
\newblock Ensemble adversarial training: Attacks and defenses.
\newblock In {\em 6th International Conference on Learning Representations,
  {ICLR} 2018, Vancouver, BC, Canada, April 30 - May 3, 2018, Conference Track
  Proceedings}. OpenReview.net, 2018.

\bibitem{XuW}
Weilin Xu, David Evans, and Yanjun Qi.
\newblock Feature squeezing: Detecting adversarial examples in deep neural
  networks.
\newblock In {\em 25th Annual Network and Distributed System Security
  Symposium, {NDSS} 2018, San Diego, California, USA, February 18-21, 2018}.
  The Internet Society, 2018.

\bibitem{XuX}
Xing Xu, Jingran Zhang, Yujie Li, Yichuan Wang, Yang Yang, and Heng~Tao Shen.
\newblock Adversarial attack against urban scene segmentation for autonomous
  vehicles.
\newblock {\em {IEEE} Trans. Ind. Informatics}, 17(6):4117--4126, 2021.

\bibitem{Yahya}
Zakia Yahya, Muhammad Hassan, Muhammad~Shahzad Younis, and Muhammad Shafique.
\newblock Probabilistic analysis of targeted attacks using transform-domain
  adversarial examples.
\newblock {\em {IEEE} Access}, 8:33855--33869, 2020.

\bibitem{YangP}
Puyudi Yang, Jianbo Chen, Cho{-}Jui Hsieh, Jane{-}Ling Wang, and Michael~I.
  Jordan.
\newblock {ML-LOO:} detecting adversarial examples with feature attribution.
\newblock In {\em The Thirty-Fourth {AAAI} Conference on Artificial
  Intelligence, {AAAI} 2020, The Thirty-Second Innovative Applications of
  Artificial Intelligence Conference, {IAAI} 2020, The Tenth {AAAI} Symposium
  on Educational Advances in Artificial Intelligence, {EAAI} 2020, New York,
  NY, USA, February 7-12, 2020}, pages 6639--6647. {AAAI} Press, 2020.

\bibitem{YuanX}
Xiaoyong Yuan, Pan He, Qile Zhu, and Xiaolin Li.
\newblock Adversarial examples: Attacks and defenses for deep learning.
\newblock {\em {IEEE} Trans. Neural Networks Learn. Syst.}, 30(9):2805--2824,
  2019.

\end{thebibliography}
}

\end{document}